\documentclass{article}
\usepackage[utf8]{inputenc}
\usepackage{graphicx}
\usepackage{algorithm}
\usepackage{algpseudocode}
\usepackage{comment}
\usepackage{enumitem}
\usepackage{makecell}
\usepackage{siunitx}
\usepackage{colortbl}
\usepackage[numbers]{natbib}
\usepackage{array}
\usepackage{amsmath,amssymb,amsfonts}

\usepackage{fancyhdr}
\title{Deep Reinforcement Learning with Interactive Feedback in a Human-Robot Environment}

\author{
Ithan Moreira$^{1,\ddagger}$ \and 
Javier Rivas$^{1,\ddagger}$ \and 
Francisco Cruz$^{1,2}$ \and 
Richard Dazeley$^{2}$ \and 
Angel Ayala$^{3}$ \and 
Bruno Fernandes$^{3}$}

\date{\normalsize
$^{1}$ Escuela de Ingenier\'ia, Universidad Central de Chile, Santiago, Chile.\\
$^{2}$ School of Information Technology, Deakin University, Geelong, Australia.\\
$^{3}$ Escola Polit\'ecnica de Pernambuco, Universidade de Pernambuco,\\Recife, Brasil.\\
Corresponding e-mails: \{ithan.moreira, javier.rivas\}@alumnos.ucentral.cl, \{francisco.cruz, richard.dazeley\}@deakin.edu.au, \{aaam, bjtf\}@ecomp.poli.br \\
$^{\ddagger}$ These authors contributed equally to this work.
}

\begin{document}

\maketitle

\begin{abstract}
Robots are extending their presence in domestic environments every day, being more common to see them carrying out tasks in home scenarios.
In the future, robots are expected to increasingly perform more complex tasks and, therefore, be able to acquire experience from different sources as quickly as possible. 
A plausible approach to address this issue is interactive feedback, where a trainer advises a learner on which actions should be taken from specific states to speed up the learning process. 
Moreover, deep reinforcement learning has been recently widely used in robotics to learn the environment and acquire new skills autonomously.
However, an open issue when using deep reinforcement learning is the excessive time needed to learn a task from raw input images.
In this work, we propose a deep reinforcement learning approach with interactive feedback to learn a domestic task in a human-robot scenario.
We compare three different learning methods using a simulated robotic arm for the task of organizing different objects; the proposed methods are (i) deep reinforcement learning (DeepRL); (ii) interactive deep reinforcement learning using a previously trained artificial agent as an advisor (agent-IDeepRL); and (iii) interactive deep reinforcement learning using a human advisor (human-IDeepRL).
We demonstrate that interactive approaches provide advantages for the learning process.
The obtained results show that a learner agent, using either agent-IDeepRL or human-IDeepRL, completes the given task earlier and has fewer mistakes compared to the autonomous DeepRL approach.
\end{abstract}

\thispagestyle{fancy}


\section{Introduction}
Robotics has been getting more attention since new researched advances have introduced significant improvements to our society.
For instance, for many years, robots have been installed in the automotive industrial area \cite{shepherd2014kuka}.
However, the current technological progress has allowed expanding the robot's applications domain in areas such as medicine, military, search and rescue, and entertainment. 
In this regard, under current research, another challenging application of robotics is its integration to domestic environments, mainly due to the presence of many dynamic variables in comparison to industrial contexts \cite{cruz2018action}.
Moreover, in domestic environments, it is expected that humans regularly interact with robots and that the robots can understand and respond accordingly to the interactions~\cite{goodrich2008human, churamani2020icub}.

Algorithms such as reinforcement learning (RL) \cite{sutton2018reinforcement} allow a robotic agent to autonomously learn new skills, in order to solve complex tasks inspired by the way as humans do, through trial and error~\cite{niv2009reinforcement}.
RL agents interact with the environment in order to find an appropriate policy that meets the problem aims.
To find the appropriate policy, the agent interacts with the environment by performing an action $a_t$ and, in turn, the environment returns a new state $s_{t+1}$ with a reward $r_{t+1}$ for the performed action to adjust the policy.
However, an open issue in RL algorithms is the time and the resources required to achieve good learning outcomes~\cite{cruz2018multi, bignold2020conceptual}, which is especially critical in online environments~\cite{ayala2019reinforcement, millan2019human}.
One of the reasons for this problem is that the agent, at the beginning of the learning process, does not know the environment and the interactions responses.
Thus, to address this problem, the agent must explore multiple paths to refine its knowledge about the environment.


In continuous spaces, an alternative is to recognize the agent's state directly from raw inputs.
Deep reinforcement learning (DeepRL) \cite{barros2020moody} is based on the same RL structure but also adds deep learning to process the function approximation for the state in multiple abstraction levels.
An example of DeepRL implementations is by convolutional neural networks (CNN) \cite{lecun2015deep} which can be modified to be used for DeepRL, e.g., DQN \cite{mnih2015human}.
CNNs have brought significant progress in the last years in different areas such as image, video, audio and speech processing, among others~\cite{krizhevsky2012imagenet}.
Nevertheless, for a robotic agent working in highly dynamic environments, DeepRL still needs excessive time to learn a new task properly.

\section{Related works}
In this section, we review previously developed works considering two main areas.
First, we address the deep reinforcement learning approach and the use of interactive feedback.
Following, we discuss the problem of vision-based object sorting using robots in order to contextualize our approach properly.
Finally, we describe the scientific contributions of this work.


\subsection{Deep reinforcement learning and interactive feedback}

Deep reinforcement learning (DeepRL) combines reinforcement Learning (RL) and deep neural networks (DNN).
This combination has allowed the enhancement of RL agents when autonomously exploring a given scenario~\cite{van2016deep}. 
If an RL agent is learning a task, the environment gives the agent the necessary information on how good or bad the taken actions are. 
With this information, the agent must differentiate which actions lead to a better accomplishment of the task aims~\cite{sutton2018reinforcement}.

Aims may be expressed by a reward function that assigns a numerical value to each performed action from a given state.
Additionally, after performing an action, the agent reaches a new state.
Therefore, the agent associates states with actions to maximize  $r_0 + \gamma \cdot r_1 + \gamma^2 \cdot r_2 + ... $, where $r_i$ is the obtained reward in the $i$-th episode and $\gamma$ is the discount factor parameter that indicates how influential future actions are. 

The fundamental base of RL tasks is the Markov decision process (MDP), in which future transitions and rewards are affected only by the current state and the selected action \cite{puterman1994markov}. 
Therefore, if the Markovian property is present for a state, this state contains all the information needed about the dynamics of a task. 
For instance, chess is a classic example of the Markov property.
In this game, it does not matter the history of plays in order to make a decision about the next movement.
All the information is already described in the current distribution of pieces over the board. 
In other words, if the current state is known, the previous transitions that led the agent to that situation become irrelevant in terms of the decision-making problem.

Formally, an MDP is defined by a 4-tuple $<S, A, \delta, r>$ where:

\begin{itemize}
\item[--] $S$ is a finite set of system states, $s \in S$;
\item[--] $A$ is a finite set of actions, $a \in A$, and $A_{s_{t}} \in A$ is a finite set of actions available in $s_t \in S$ at time $t$;
\item[--] $\delta$ is the transition function $\delta:S \times A \rightarrow S$;
\item[--] $r$ is the immediate reward (or reinforcement) function $r:S \times A \rightarrow \mathbb{R}$.
\end{itemize}

Each time-step $t$, the agent perceives the current state $s_t \in S$ and chooses an action $a_t \in A$ to be carried out. 
The environment gives a reward $r_t= r(s_t, a_t)$ and the agent moves to the state $s_{t+1}= \delta(s_t, a_t)$. 
The functions $r$ and $\delta$ depend only on the current state $s_t$ and action $a_t$, therefore, it is a no memory process.
Over time, the agent tries to learn a policy $\pi: S \rightarrow A$ which, from a state $s_t$, yields the greatest value or discounted reward \cite{sutton2018reinforcement}, as shown in Eq. \eqref{Eq:value}.

\begin{equation}
Q^{\pi}(s_t,a_t) = r_t+ \gamma \cdot r_{t+1} + \gamma^2 \cdot r_{t+2} + ... = \sum_{i=0}^{\infty} \gamma^i \cdot r_{t+1}
\label{Eq:value}
\end{equation}
where $Q^{\pi}(s_t,a_t)$ is the action-value function following the policy $\pi$ (e.g., choosing action $a_t$) from a given state $s_t$.
The discount factor $\gamma$ is a constant ($0 \leq \gamma < 1$) which determines the relative importance of immediate and future rewards. 
For instance, in case $\gamma = 0$, then the agent is short-sighted and maximizes only the immediate rewards, or in case $\gamma \rightarrow 1$ the agent is more foresighted in terms of future reward.

The final goal of RL is to find an optimal policy ($\pi^*$) mapping states to actions in order to maximize the future reward ($r$) over a time ($t$) with a discount rate ($\gamma \in[0,1]$), as shown in Eq. \eqref{eq:action-value}.
In the equation, $\mathbb{E}_{\pi}[ ]$ denotes the expected value given that the agent follows policy $\pi$ and $Q^{*}(s_t,a_t)$ denotes the optimal action-value function~\cite{sutton2018reinforcement}. 
Table \ref{Table:elements} summarizes all the elements shown within Eq. \eqref{eq:action-value}. 
In DeepRL, an approximation function, implemented by DNN, allows an agent to work with high-dimensional observation spaces, such as pixels of an image~\cite{mnih2015human}.


\begin{equation}
    Q^{*}(s_t,a_t)= max \left( \mathbb{E}_{\pi}[r_{t}+\gamma r_{t+1}+\gamma^{2}  r_{t+2}+...|s_t=s,a_t=a,\pi] \right)
    \label{eq:action-value}
\end{equation}

\begin{table}[H]
\caption{Elements to compute the optimal action-value function.}
\centering
\begin{tabular}{ll}
\hline
\textbf{Symbol}	& \textbf{Meaning}\\
\hline
$Q^{*}(s,a)$    & Optimal action-value function\\
$\mathbb{E}$    & Expected value following policy $\pi$\\
$\pi$		    & Policy to map states to actions\\
$\gamma$	    & Discount factor\\
$r_t$		    & Reward received at time step $t$\\
$s_t$		    & Agent's state at time step $t$\\
$a_t$		    & Action taken at time step $t$\\
\hline
\end{tabular} \label{Table:elements}
\end{table}

Interactive feedback is a method that improves the learning time of an RL agent \cite{suay2011effect}. 
In this method, an external trainer can guide the agent's apprenticeship to explore more promising areas at early learning stages.
The external trainer is an agent that can be a human, a robot, or another artificial agent.

\begin{figure}
    \centering
    \includegraphics[width=0.8\textwidth]{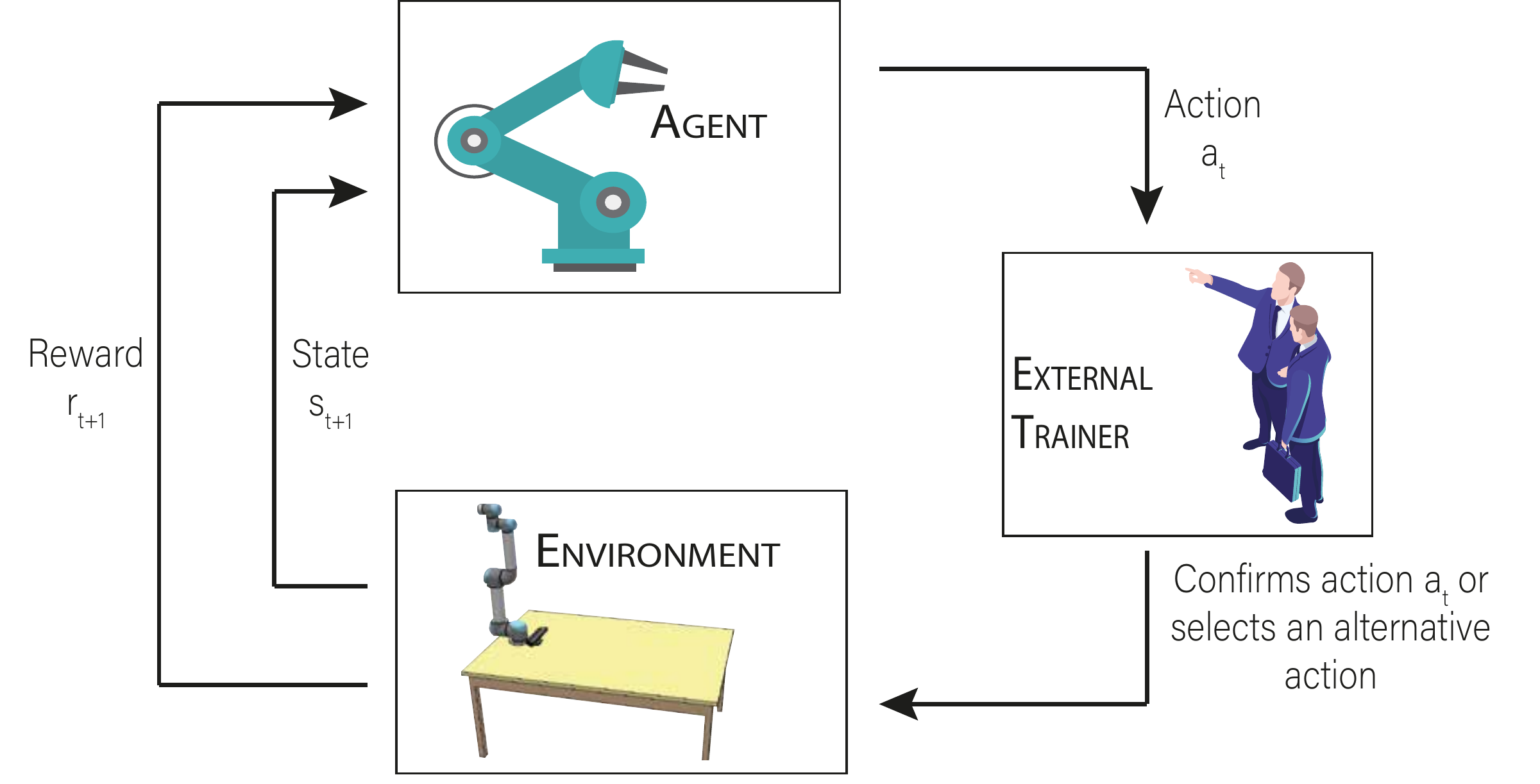}
    \caption{Policy-shaping interactive feedback approach. In this approach, the trainer may advise the agent on what actions to take in a particular given state.} 
    \label{fig:IDeepRL}
\end{figure}

There are two principal strategies for providing interactive feedback in RL scenarios, i.e., evaluative and corrective feedback \cite{najar2020reinforcement}.
In the first one, called reward-shaping, the trainer modifies or accepts the reward given by the environment in order to bias the agent's learning~\cite{ng1999policy, brys2014combining}.
In the second one, called policy-shaping, the trainer may suggest a different action to perform, by replacing the one proposes by the policy \cite{griffith2013policy, li2019human}.
A simple policy-shaping method involves forcing the agent to take certain actions that are recommended by the trainer~\cite{grizou2013robot, navidi2020human}.
For instance, a similar approach is used when a teacher is guiding a child's hand to learn how to draw a geometric figure.
In this work, we use the policy-shaping approach since it has been shown that humans using this technique to instruct an agent provide advice that is more accurate, are able to assist the learner agent for a longer time, and provide more advice per episode.
Moreover, people using policy-shaping have reported that the agent's ability to follow the advice is higher, and therefore, felt their own advice to be of higher accuracy when compared to people providing advice via reward-shaping \cite{bignold2019rule}.
The policy-shaping approach is depicted in Figure \ref{fig:IDeepRL}.

There are different ways to support the agent's learning, which in turn may lead to other problems. 
For instance, if the trainer delivers too much advice, the learner never gets to know other alternatives because most of the decisions taken are given from the external trainer \cite{taylor2014reinforcement}.
The quality of the given advice by the external trainer must also be considered to improve the learning.
It has been shown that inconsistent advice may be very detrimental during the learning process, so that in case of low consistency of advice, autonomous learning may lead to better performance~\cite{cruz2018improving}.

One additional strategy to better distribute the given advice is to use a budget \cite{taylor2014reinforcement}.
In this strategy, the trainer has a limited amount of interaction with the learner agent, similar to the limited patience of a person for teaching.
There are different ways of using the budget, in terms of when to interact or give advice, namely, early advising, alternating advice, importance advising, mistake correcting, and predictive advising.
In this work, we use early advising allowing us to fairly compare interactive approaches using the different kinds of trainers used in the proposed methods, i.e., humans or artificial agents as trainers.

Although there have been some approaches addressing the interactive deep reinforcement learning problem, they have been mostly used in other scenarios.
For instance, in \cite{dobrovsky2019improving} is presented an application to serious games, and in \cite{rajeswaran2017learning} is presented a dexterous robotic manipulation approach using demonstrations. 
In the game scenario \cite{dobrovsky2019improving}, the addressed task presents different environmental dynamics compared to human-robot environments.
Moreover, the authors propose undoing an action by the advisor, which is not always feasible.
For example, in a human-robot environment, a robot might break an object as a result of a performed action, which is impossible to undo.

\subsection{Vision-based object sorting with robots}

The automation of sorting object tasks has been previously addressed using machine learning techniques.
For instance, Lukka et al. \cite{lukka2014zenrobotics} implemented a recycling robot for construction and demolition waste.
In this work, the robot sorts the waste using a vision-based system for object recognition and object manipulation to control the movement of the robot in order to classify the objects presented on a moving belt properly.
The authors did not present performance results since the approach is presented as a functional industrial prototype for sorting objects through images.

The object recognition problem is an extended research area that has been addressed by different techniques, including deep learning, as presented in  \cite{zhihong2017robotgrasping}.
This approach is a similar system to \cite{lukka2014zenrobotics} in terms of proposing to sort garbage from a moving belt.
The authors used a convolutional neural network, called Fast R-CNN, to obtain the moving object's class and location, and send the information to the robotic grasping control unit to grasp the object and move it.
As the authors point out, the key problem the deep learning method tries to solve is the object identification.
Moreover, another approach to improve the object recognition task is presented in \cite{sun2017clothing}, where the authors implemented a stereo vision system to recognize the material and the clothes categories.
The stereo vision system creates a 3D reconstruction of the image to process and obtains local and global features to predict the clothing category class and manipulate a robot that must grasp and sort the clothing in a preestablished box.
These two systems, presented in \cite{zhihong2017robotgrasping} and \cite{sun2017clothing}, use a supervised learning method that requires prior training of items to be sorted, leading to low generalization for new objects.

Using RL to sort objects has also been previously addressed.
For instance, in \cite{cruz2014improving} a cleaning-table task in a simulated robotic scenario is presented.
In order to complete the task, the robot needs to deal with objects such as a cup and a sponge.
In this work, the RL agent used a discrete tabular RL approach complemented by interactive feedback and affordances \cite{cruz2016learning}. 
Therefore, the agent did not deal with the problem of continuous visual inputs for state representation.
Furthermore, in \cite{zhang2015towards} an approach for robotic control using DeepRL is presented.
In this work, a simulated Baxter robot learned autonomous control using a DQN-based system.
When transferring the system to a real-world scenario, the approach failed. 
To fix this, the system ran replacing the camera images with synthetic images in order for the agent to acquire the state and decide which action to take in real-world.


\subsection{Scientific contribution}
Although a robot may be capable of learning autonomously to sort objects in different contexts, current approaches address the problem using supervised deep learning methods with previously labeled data to recognize the objects, e.g., \cite{zhihong2017robotgrasping} and \cite{sun2017clothing}.
In this regard, using the DeepRL approach allows classifying objects as well as deciding how to act with them.
Additionally, if prior knowledge of the problem is transferred to the DeepRL agent, e.g., using demonstrations \cite{vecerik2017leveraging}, the learning speed may also be improved.
Therefore, using interactive feedback as an assistance method, we will be able to advise the learner agent during the learning process, using both artificial and human trainers, to evaluate how the DeepRL method responds to the given advice.

In this work, we present an interactive-shaping vision-based algorithm derived from DeepRL, referred to here as interactive DeepRL or IDeepRL.
Our algorithm allows us to speed up the required learning time through strategic interaction with either a human or an artificial advisor.
The information exchange between the learner and the advisor gives the learner a better understanding of the environment by reducing the search space.

We have implemented a simulated domestic scenario, in which a robot has to organize objects considering color and shape.
The RL agent perceives the world through RGB images and interacts through a robotic arm while an external trainer may advise the agent a different action to perform during the first training steps.
The implemented scenario is used for test and comparison between the DeepRL and IDeepRL algorithms, as well as the evaluation of IDeepRL using two different types of trainers, namely, another artificial agent previously trained and a human advisor.

Therefore, the contribution of the proposed method is to demonstrate that interactive-shaping advice can be efficiently integrated into vision-based deep learning algorithms. 
The interactive learning methodologies proposed in this work outperform current autonomous DeepRL approaches allowing to collect more and faster reward using both artificial and human trainers.

\section{Material and methods} 

\subsection{Methodology and implementation of the agents} \label{section:design}
In this work, our focus is on assessing how interactive feedback, used as an assistance method, may affect the performance of a DeepRL agent. 
To this aim, we implement three different approaches for the RL agents:

\begin{enumerate}[label=\roman*.]
\item DeepRL: where the agent interacts autonomously with the environment; 
\item agent-IDeepRL: where the DeepRL approach is complemented with a previously trained artificial agent to give advice; and 
\item human-IDeepRL: where the DeepRL approach is complemented with a human trainer.
\end{enumerate}

The first approach includes a standard deep reinforcement learning agent, referred to here as DeepRL, and is the basis of both of the interactive agents discussed subsequently. 
The DeepRL agent perceives the environment information through a visual representation~\cite{desai2017deep}, which is processed by a convolutional neural network (CNN) that estimates the Q-values.
The deep Q-learning algorithm allows the agents to learn through actions previously experienced, using the CNN as a function approximator, allowing them to generalize states and apply Q-learning in continuous state spaces.

To save past experiences, the experience replay \cite{adam2012experience} technique is implemented. 
This technique saves the most useful information (experience) in memory, which is used afterward to train the RL agent. 
The neural network is responsible for processing the visual representation and gives the Q-value of each action to the agent, which decides what action to take.
In order for the agent to balance exploration and exploitation of actions, the $\epsilon$-greedy method is used.
This method includes an $\epsilon$ parameter, which allows the agent to performs either a random exploratory action or the best known action proposed by the policy.

The learning process for the autonomous agent, i.e., DeepRL agent, is separated into two stages.
The first pretraining stage consists of 1000 random actions that the agent must perform to populate the initial memory.
In the second stage, the agent's training is carried out using the $\epsilon$-greedy policy and, after each performed action, the agent is trained using 128 tuples considering state, action, reward, and next state, as $<s_t, a_t, r_t, s_{t+1}>$ extracted from the memory.

\begin{figure}
    \centering
    \includegraphics[width=0.9\textwidth]{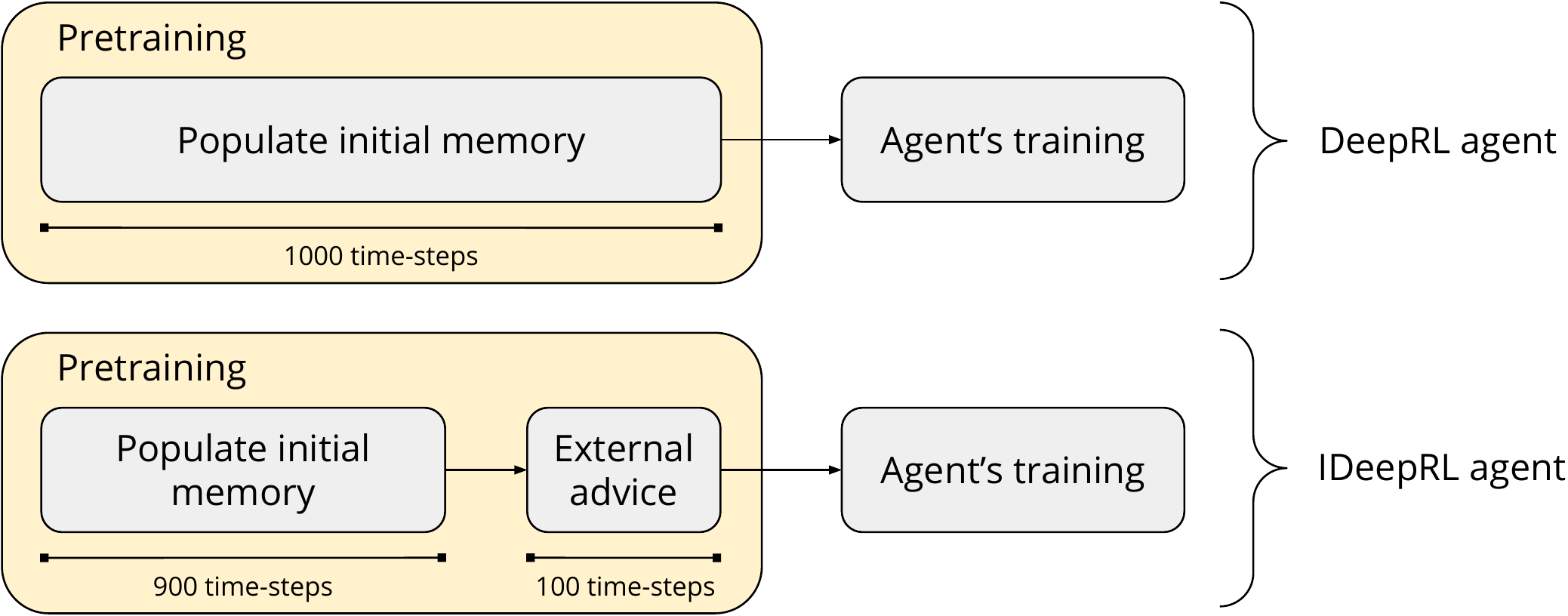}
    \caption{The learning process for autonomous and interactive agents. Both approaches include a pretraining stage comprising 1000 actions. For interactive agents, the final part of the pretraining is performed using external advice instead of random actions.}
    \label{fig:Pretraining}
\end{figure}

Both IDeepRL approaches are based on autonomous DeepRL and include the interactive feedback strategy from an external trainer to improve the DeepRL performance.
Therefore, the agents have the same base algorithm, adding an extra interactive stage.
For the interactive agents, the learning process is separated into three stages.
The first pretraining stage corresponds to 900 random actions that the agent must perform in order to populate the initial memory.
In the second stage, the external trainer participates giving early advice about the environment dynamics and the intended task during 100 consecutive time-steps. 
In the third stage, the agent starts training using the $\epsilon$-greedy policy, and, following each action selected, the agent is trained with 128 tuples as $<s_t, a_t, r_t, s_{t+1}>$ saved previously in the batch memory. 
The learning process for both autonomous and interactive agents is depicted in Figure \ref{fig:Pretraining}.

\begin{algorithm}[t]
\begin{algorithmic}[1]
\State Initialize memory M
\State Observe agent's initial state $s_0$
\While{len(M) $\leq 1000$}
  \If {interaction is used AND length of M $> 900$} \label{Lin:interactive}
    \State Get action $a_t$ from advisor
  \Else
    \State Choose a random action $a_t$
  \EndIf
  \State Perform action $a_t$
  \State Observe $r_{t}$ and next state $s_{t+1}$
  \State Add ($<s_t, a_t, r_t, s_{t+1}>$) to M
  \If {$s_t$ is terminal OR time-steps $> 250$} 
    \State Reset episode
  \EndIf
\EndWhile
\end{algorithmic}
\caption{Pretraining algorithm to populate the batch memory including interactive feedback.} \label{alg:Pretraining}
\end{algorithm}

\begin{algorithm}
\begin{algorithmic}[1]
\State Perform the pretraining Algorithm 1
\For{each episode}
  \State Observe state $s_t$
  \Repeat
    \State Choose an action $a_t$ using $\epsilon$-greedy 
    \State Perform action $a_t$
    \State Observe $r_{t}$ and next state $s_{t+1}$
    \State Add ($<s_t, a_t, r_t, s_{t+1}>$) to M
    \State Populate randomly batch B from M
    \State Train CNN using data in B
    \State $s_{t} \leftarrow s_{t+1}$
    \State $\epsilon \leftarrow \epsilon * \epsilon$\_decay
  \Until{$s_t$ is terminal OR time-steps $> 250$}
\EndFor
\end{algorithmic}
\caption{Training algorithm to populate and extract information from the batch memory.} \label{alg:Training}
\end{algorithm}

In the second stage of the IDeepRL approach, the learner agent receives advice either from a previously trained artificial agent or from a human trainer.
The artificial trainer agent used in agent-IDeepRL is an RL agent that collected previous experience by performing the autonomous DeepRL approach using the same hyperparameters. Therefore, the knowledge is acquired by interacting with the environment in the same manner as the learner agent does.
Once the trainer agent has learned the environment, it is used to then provide advice in agent-IDeepRL over 100 consecutive time-steps. 
Both the trainer agent and the learner agent perceived the environmental information through a visual representation after an action is performed.

Algorithm \ref{alg:Pretraining} shows the first stage for DeepRL and IDeepRL approaches, which corresponds to the pretraining stages to populate the batch memory.
This algorithm also contains the second stage for interactive agents using IDeepRL represented with a conditional in line \ref{Lin:interactive}.
Moreover, in Algorithm \ref{alg:Training} is observed the second stage for DeepRL, which also corresponds to the third stage for IDeepRL, the training stage.

\subsubsection{Interactive approach}

As previously discuss, the IDeepRL methods include an external advisor, which can be another already trained agent or human. 
In our scenario, the advisor uses a policy-shaping approach during the decision-making process, as previously shown in Figure \ref{fig:IDeepRL}. 
Moreover, between the different alternatives for interactive feedback, we use teaching on a budget with early advising \cite{cruz2017agent}.
This technique attempts to reduce the time required for an RL agent to understand better the environment, achieved by 100 early consecutive pieces of advice from the trainer, trying to transfer the trainer's knowledge of the environment as quickly as possible.
For the different ways to implement the interactive approach, we use early advising for the training of the learner agent, using a limited consecutive amount of advice to be used by the trainer to help the agent.

\subsubsection{Visual representation}

A visual representation for the deep Q-learning algorithm is used, which consists of Q-learning using a function approximator for the Q-values with a CNN.
Additionally, it uses a memory with past experiences from where are taken batches for the network training.

\begin{figure}[t]
    \centering
    \includegraphics[width=1.0\textwidth]{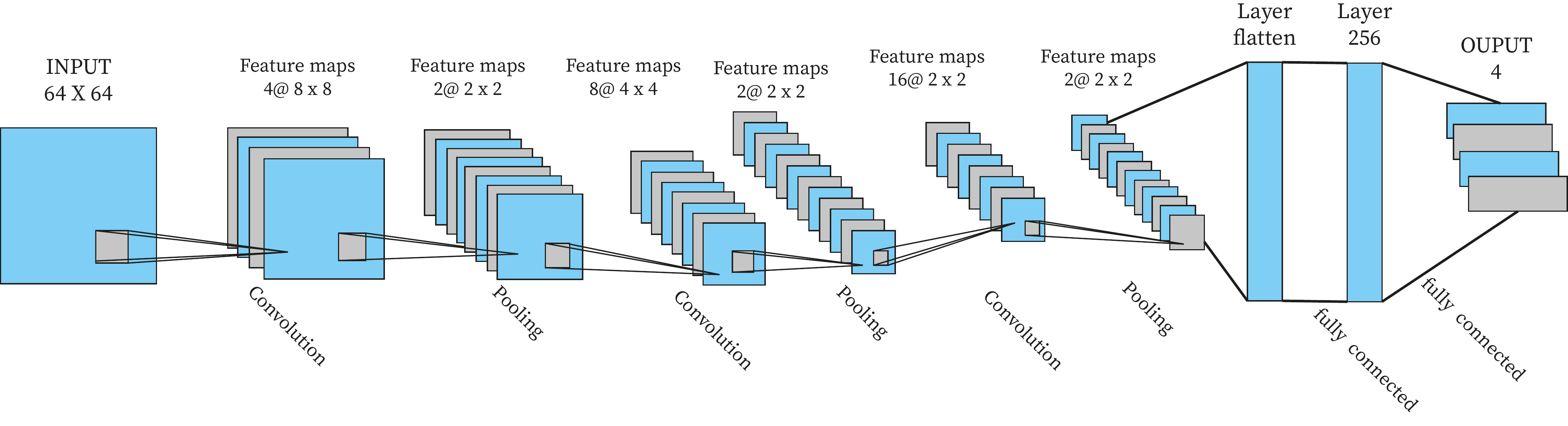}
    \caption{Neural network architecture, with an input of a $64 \times 64$ RGB image, and composed of three convolution layers, three max-pooling layers, and two fully connected layers, including a softmax function for the output.}
    \label{fig:NeuralNetworkArchitecture}
\end{figure}

Our architecture is capable of processing input images of $64 \times 64$ pixels used in RGB channels for learning the image features.
The architecture is inspired by similar networks used in other DeepRL works \cite{mnih2015human, krizhevsky2012imagenet}.
Particularly, in the first layer, an $8 \times 8$ convolution with four filters is used, then a $2 \times 2$ max-pooling layer followed by a $4 \times 4$ convolution layer with eight filters, and followed by another max-pooling with the same specification as the previous one.
The network has a last $2 \times 2$ convolution layer with 16 filters.
After the last pooling, a flatten layer is applied, which is fully connected to a layer with 256 neurons.
Finally, the 256 neurons are also fully connected with the output layer, which uses a softmax function, including four neurons to represent the possible actions.
The full network architecture can be seen in Figure \ref{fig:NeuralNetworkArchitecture}.
Since this work is oriented to compare different learning methodologies, all agents were trained with the same architecture to compare them fairly.

\subsubsection{Continuous representation}

Given the task characteristics, considering images as inputs to recognize different objects in a dynamic environment, it is impractical to generate a table with all the possible state-action combinations. Therefore, we have used a continuous representation combining two methods.
The first method is a function approximator through a neuronal network for Q($s_t$, $a_t$), which allows us to generalize the states, in order to use Q-learning in continuous spaces and select which action is carried out. 
The second method is the experience replay technique, which saves on memory a tuple of an experience given by $<s_t, a_t, r_t, s_{t+1}>$.
These data saved in memory are used afterward to train the RL agent.

\subsection{Experimental setup}

We have designed a simulated domestic scenario focused on organizing objects.
The agent aims to classify geometric figures with different shapes and colors and organize them in designated locations to optimize the collected reward.
Classification tasks are widespread in domestic scenarios, e.g., organizing cloth.
The object shape might represent different cloth types, while the color might represent whether it is clean or dirty.

In order to compare DeepRL and IDeepRL algorithms, three different agents are trained in this scenario in terms of collected reward and learning time.
The experimental scenario is developed in the simulator CoppeliaSim developed by Coppelia Robotics \cite{rohmer2013vrep}.

\begin{figure}
    \centering
    \includegraphics[width=0.9\textwidth]{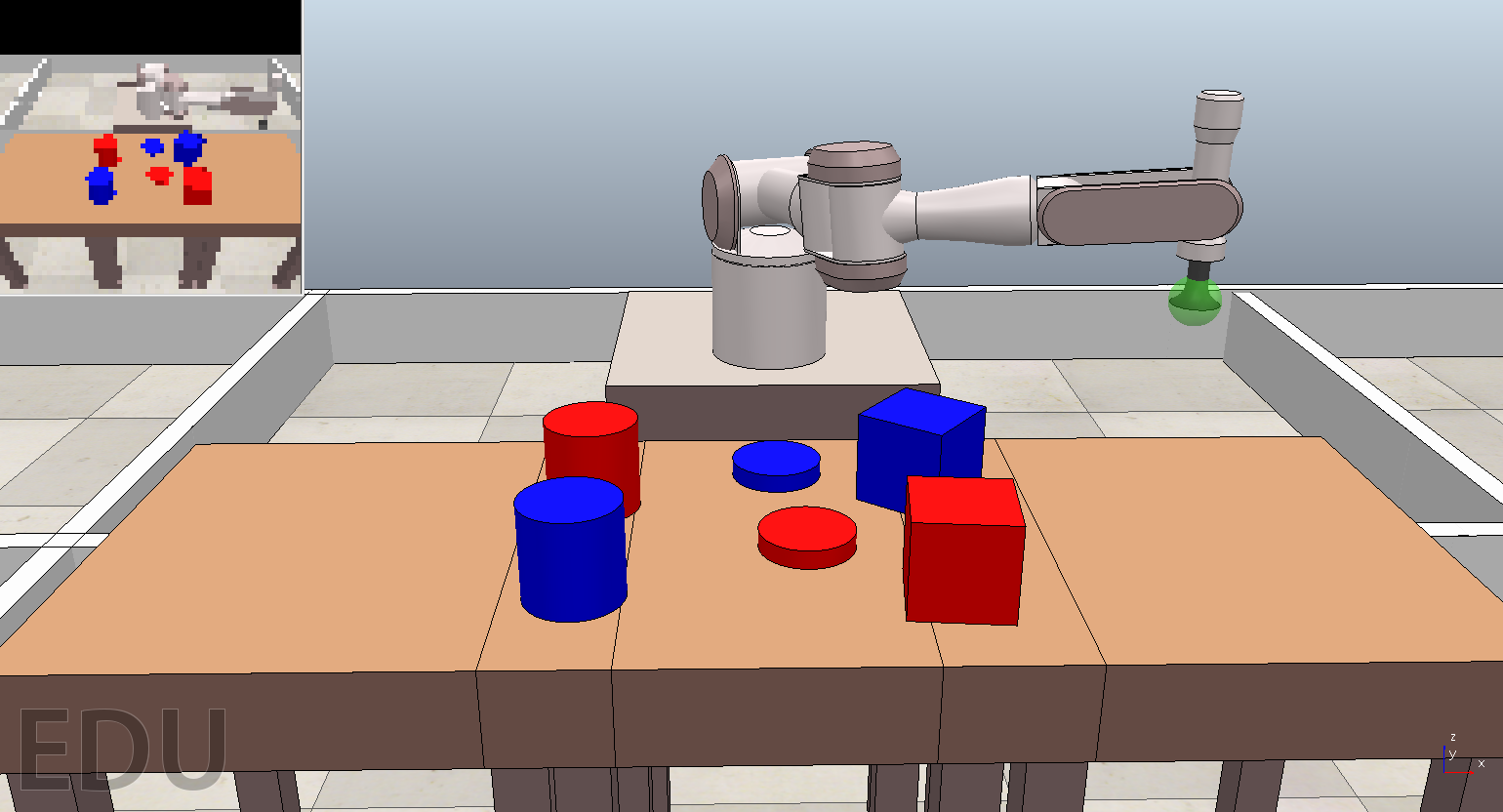}
    \caption{The simulated domestic scenario presenting 6 objects in different colors and the robotic arm.
    In the upper left corner is shown the camera signal, which is a $64 \times 64$ pixels RGB image for the state representation of the agent.}
    \label{fig:RoboticScenario}
\end{figure}

Three tables are used in the scenario; the first contains all the unsorted objects initially placed randomly on the table within nine preestablished positions. 
This represents the initial dynamic state. 
The second two tables are used to place the objects once the agent determines which table objects belong. 
To perform this sort a robotic manipulator arm with 7 degrees of freedom, six axes, and a suction cup grip is used. 
The robotic arm is placed on another table along with a camera from where we obtain RGB images.
The objects to be organized are cubes, cylinders, and disks in two different colors, red and blue, as are presented in Figure \ref{fig:RoboticScenario}.

\subsubsection{Actions and state representation}

The available actions for the agent are four and can be taken in an autonomous way or through given advice from the external trainer.
The actions are the following:

\begin{enumerate}[label=\roman*.]
    \item{Grab object:} the agent grabs randomly one of the objects with the suction cup grip.
    \item{Move right:} the robotic arm is moved to the table on the right side of the scenario; if the arm is already there, do nothing.
    \item{Move left:} the robotic arm is moved to the table on the left side of the scenario; if the arm is already there, do nothing.
    \item{Drop:} if the robotic arm has an object in the grip and is located in one of the side tables, the arm goes down and releases the object; in case the arm is positioned in the center, the arm keeps the position.
\end{enumerate} 

For example, the actions required to correctly organize a blue cube from the central table consist of (i) grab object, (ii) move right, and (iii) drop.
The robot low-level control to reach the different positions within the scenario is performed using inverse kinematics. 
Although to reach an object we use inverse kinematics, the CNN is responsible for deciding to perform the action \textit{grab an object} through the Q-values, and if so, to decide where to place the object, based on the classification.

The state comprises a high-dimensional space, represented by a raw image captured by an RGB camera.
The image presents a dimension of $64 \times 64$ pixels from where the agent perceives the environment and chooses what action to take according to the network output.
The input image is normalized to values $\in[0,1]$ to be presented to the convolutional neural network.

\subsubsection{Reward}

In terms of the reward function, there are different subtasks to complete an episode.
To complete the task successfully, all the objects must be correctly organized.
To organize one object is considered a partial goal of the task.
All the objects are initially located in the central table to be sorted, and once placed in the side tables, they cannot be grasped again.
If all the objects are correctly sorted, the reward is equal to 1, and for correctly organizing a single object, the reward is equal to 0.4.
For example, if the classification of the six objects is correct, each of the first five organized objects leads to a reward of 0.4, and the last one obtains a reward of 1, summarizing a total reward of 3. 
Furthermore, to encourage the agent to accomplish the classification task in fewer steps, a small negative reward of -0.01 per step is considered when the steps are more than 18, which is the minimal time-steps needed to complete the task satisfactorily.
If an object is misplaced, the current training episode ends, and the agent receives a negative reward of -1.
The complete reward function is shown in Eq. \eqref{Eq:reward}.

\begin{equation}
    r(s) = \left\{
    \begin{array}{r l}
          1 & \textrm{if all the objects are organized}\\
         0.4 & \textrm{if a single object is organized}\\
          -1 & \textrm{if an object is incorrectly organized}\\
         -0.01 & \textrm{if steps } > 18\\
    \end{array}
    \right.
    \label{Eq:reward}
\end{equation}

\subsubsection{Human interaction}

In the case of human trainers giving advice during the second stage of the IDeepRL approach, a brief three-step induction is carried out for each participant:

\begin{enumerate}[label=\roman*.]
    \item The user reads an introduction to the scenario and the experiment, as well as the expected results.
    \item The user is given an explanation about the problem and how it can be solved.
    \item The user is taught how to use the computer interface to advise actions to the learner agent.
\end{enumerate}

In general terms, the participants chosen for the experiment have not had significant exposure to artificial intelligence, and are not familiar with simulated robotic environments.
The solution explanation is given to the participants in order to give to all of them an equal understanding of the problem and thus to reduce the time that they spend exploring the environment and focus on advising the agent.
Each participant communicates with the learner agent using a computer interface while observing the current state and performance in the robot simulator.
The user interface contains in a menu all possible actions that can be advised.
These action possibilities are shown at the screen, and the trainer may choose any of them to be performed by the learner agent.
There is no time limit to advise each action, but, as mentioned, during the second stage of IDeepRL, the trainer has a maximum of 100 consecutive time-steps available for advice.

\vspace{1cm}

\section{Results}

In this section, we show the experimental results obtained during the training of three different agents implemented with the three proposed methodologies, i.e., an autonomous agent scenario, a human-agent scenario, and an agent-agent scenario, namely, DeepRL, human-IDeepRL, and agent-IDeepRL.
The methodologies are tested with the same hyperparameters, which have been experimentally determined concerning our scenario, as follows: initial value of $\epsilon = 1$, $\epsilon$ decay rate of 0.9995, learning rate $\alpha = 10^{-3}$, and discount factor $\gamma = 0.9$ during 300 episodes.

\begin{figure}[t]
    \centering
    \includegraphics[width=0.9\textwidth]{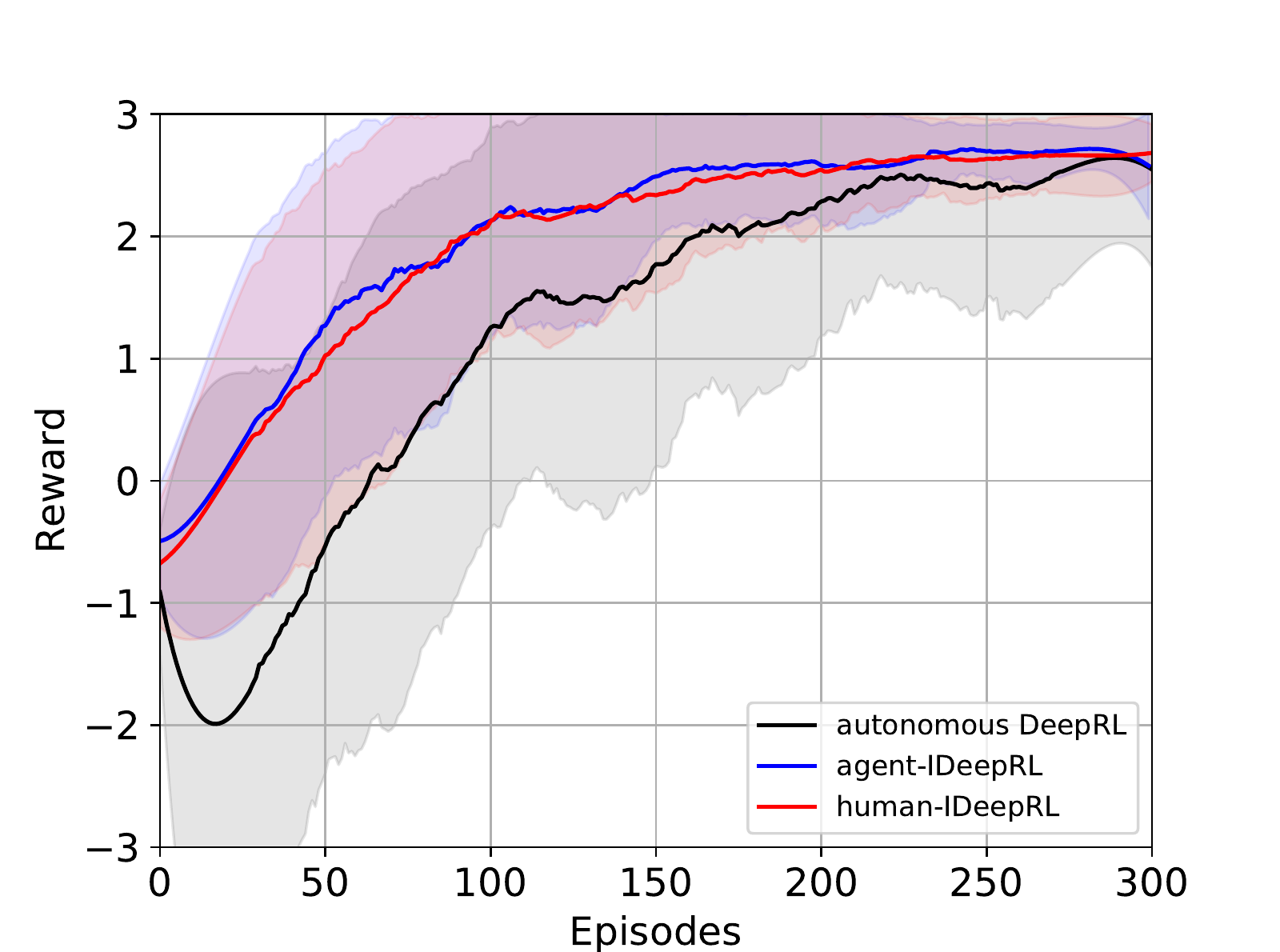}
    \caption{Average collected reward for the three proposed methods.
    The black line represents the autonomous (DeepRL) agent, which has to discover the environment without any help.
    The blue and red lines are the agents with an external trainer, namely an artificial advisor (agent-IDeepRL) and a human advisor (human-IDeepRL), respectively.
    The shadowed area around the curves shows the standard deviation for ten agents.
    The methods agent-IDeepRL and human-IDeepRL collect $64.03\%$ and $59.40\%$ more reward $R_T$ when compared to the autonomous DeepRL baseline. 
    }
    \label{fig:ResultsAutonomousvsIDeepRL}
\end{figure}

As discussed in section \ref{section:design}, the first methodology is an autonomous agent using DeepRL, who must learn how the environment works and how to complete the task. 
Given the complexity of learning the task autonomously, the time required for the learning process is rather high.
The average collected reward for ten autonomous agents is shown in Figure \ref{fig:ResultsAutonomousvsIDeepRL} represented by the black line.
Moreover, this complexity also increases the error rate or misclassification of the objects located in the central table.

Next, we perform the interactive learning approaches by using agent-IDeepRL and human-IDeepRL.
The average obtained results for ten interactive agents are shown in Figure \ref{fig:ResultsAutonomousvsIDeepRL} represented by the blue line and the red line, respectively.
The agent-IDeepRL approach performs slightly better than the human-IDeepRL approach, mainly because people needed more time to understand the setup and to react during the experiments.
However, both interactive approaches obtain very similar results, achieving much faster convergence when comparing to the autonomous DeepRL approach.
Furthermore, the learner agents getting advice from external trainers make fewer mistakes, especially at the beginning of the learning process, and are able to learn the task in fewer episodes.
On the other hand, the autonomous agent makes more mistakes at the beginning since it is trying to learn how the environment works and the aim that it has to be accomplished.
This is not the case for the interactive agents since the advisors help them during this critical part of the learning.
For a qualitative analysis of the obtained results, we have computed the total collected reward $R_T$ defined as the sum of all individual rewards $r_i$ received during the learning process, i.e., $N=300$ for this case (see Eq. \eqref{Eq:TotalReward}).
The total collected rewards for the three methods are: autonomous DeepRL $R_T=369.8788$, agent-IDeepRL $R_T=606.6995$, and human-IDeepRL $R_T=589.5974$.
Therefore, the methods agent-IDeepRL and human-IDeepRL present an improvement, in terms of collected reward, of $64.03\%$ and $59.40\%$ respectively, in comparison to the autonomous DeepRL used as a baseline.

\begin{equation}
    R_T = \sum_{i}^{N} r_i
    \label{Eq:TotalReward}
\end{equation}

Due to the trainer has a budget of 100 actions to advise, the interactive feedback is consumed within the first six episodes, taking into account that the minimal amount of actions to complete an episode are 18 actions.
Even with such a small amount of feedback, the learner agent receives an important knowledge from the advisor that is complemented with the experience replay method.
During the human-IDeepRL approach, to give demonstrative advice, 11 people participated in the experiments, four males and seven females, with ages between 16 and 24 ($M=21.63, SD=2.16$).
The participants were explained how to help the robot to complete the task giving advice using the same script for all of them (see Section \ref{section:design}). 

\begin{figure}
    \centering
    \includegraphics[width=0.9\textwidth]{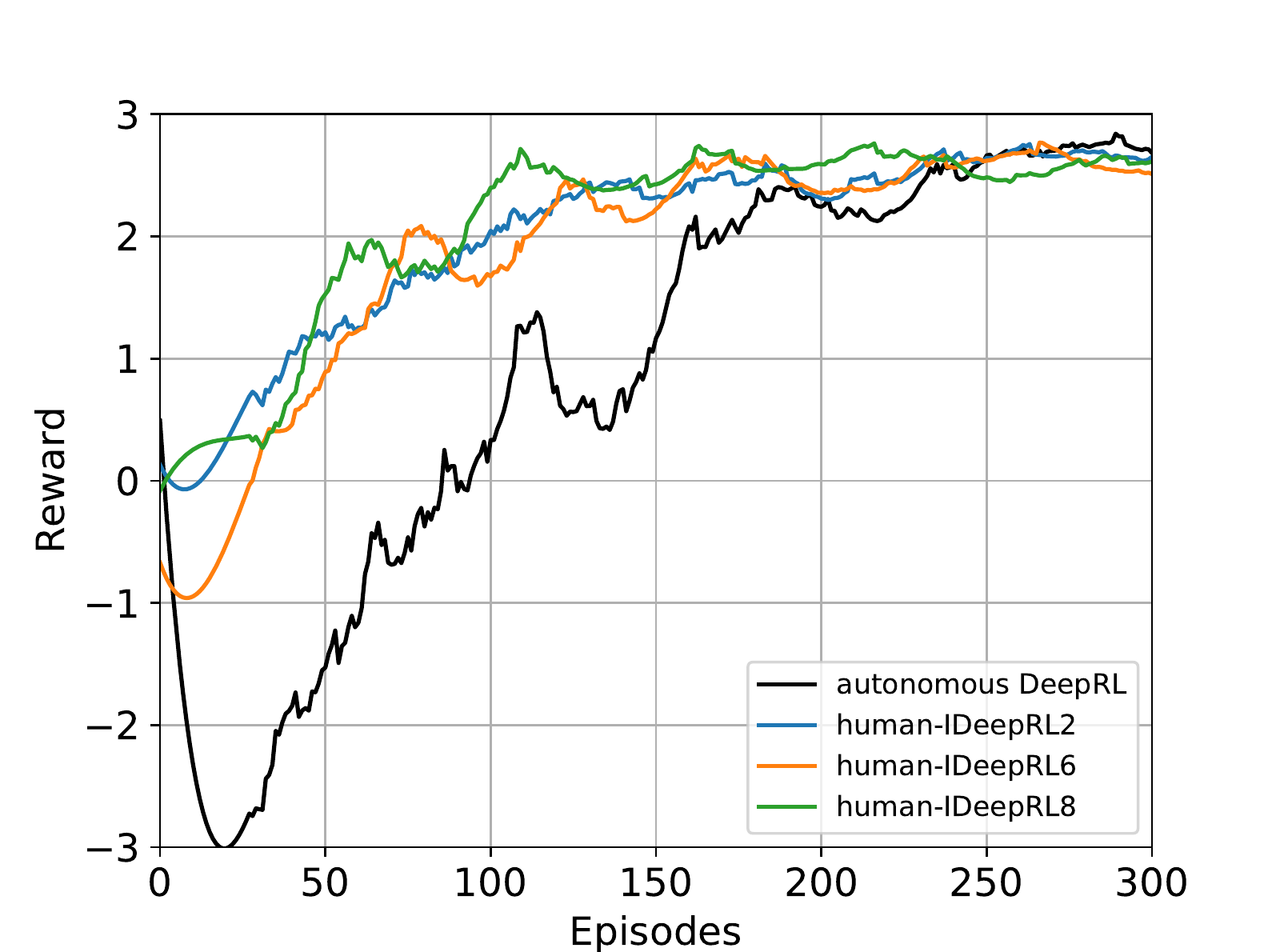}
    \caption{
    Collected rewards for a selection of example interactive agents. 
    The figure compares the learning process of agents trained by different people using human-IDeepRL (the black line is an autonomous agent that is included as a reference).
    The three human examples differ from each other by following different strategies to teach the learner agent, especially at the beginning. 
    For example, the sixth trainer (yellow) started giving wrong advice, leading to less reward collected initially, while the eighth trainer (green) started giving much better advice, which is reflected in the accumulated reward at the beginning, however, it experienced a drop in the collected reward some episodes later. 
    Although each person has initially a different understanding of the environment considering objectives and possible movements, all the interactive agents converge to similar behavior at the end of the learning process.
    Quantitatively, the interactive agents collected $126.85\%$, $110.99\%$, and $136.94\%$ more reward $R_T$ when compared to the autonomous DeepRL agent.
} 
    \label{fig:ResultsHumanIDeepRL}
\end{figure}

In Figure \ref{fig:ResultsHumanIDeepRL} are shown the collected rewards by an autonomous DeepRL agent and three learner agents trained by different people as examples.
All human-IDeepRL approaches work much better than the autonomous DeepRL, making fewer mistakes and, therefore, collecting faster rewards.
Between the interactive agents, there are some differences, especially at the beginning of the learning process, within the first 50 episodes.
It is possible to observe the different strategies followed by the human trainers, for instance, the sixth trainer (yellow line, labeled as human-IDeepRL6) started giving wrong advice, leading to less reward at the beginning, while the eighth trainer (green line, labeled as human-IDeepRL8) started giving almost perfect advice, experiencing a drop in the collected reward some episodes later. 
Nevertheless, all the agents, even the autonomous, managed well the task reaching a similar reward. 
As in the previous case, we have computed the total collected reward $R_T$ (see Eq. \eqref{Eq:TotalReward}) for each agent shown in Figure \ref{fig:ResultsHumanIDeepRL}.
The autonomous DeepRL agent collected 266.0905 of total reward, whereas the agents with the human-IDeepRL method using as trainers subjects 2, 6, and 8 collected 603.6375, 561.4384, and 630.4684 respectively.
Although there are differences in the way that each trainer instructs the learner agent, yet they have accomplished an improvement in the total accumulated reward of $126.85\%$, $110.99\%$, and $136.94\%$.

\begin{figure}[t]
    \centering
    \includegraphics[width=0.9\textwidth]{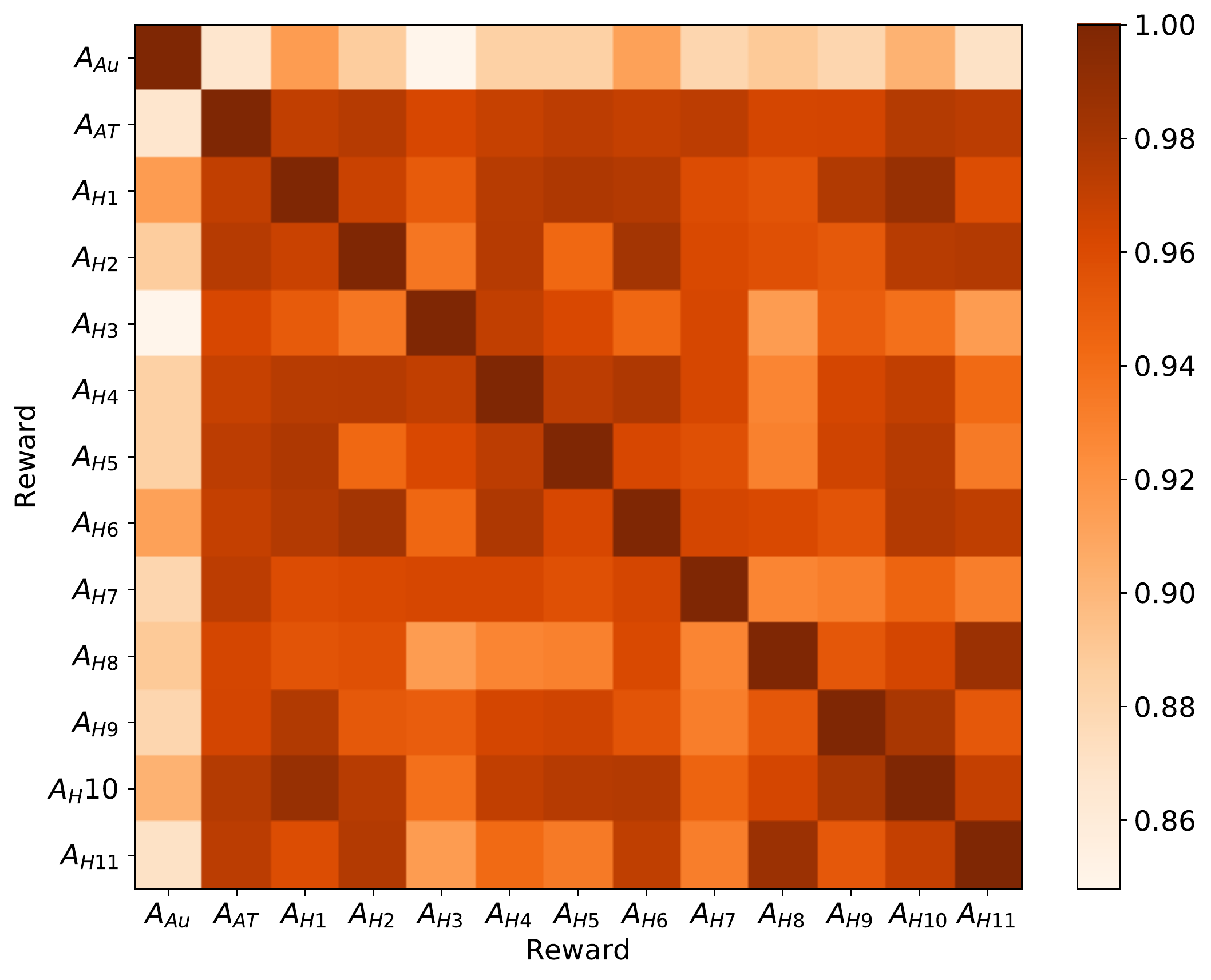}
    \caption{Pearson’s correlation between the collected rewards for different agents. 
    The shown agents include an autonomous agent ($A_{Au}$), an interactive agent trained by an artificial trainer ($A_{AT}$), and the interactive agents trained by humans (from $A_{H1}$ to $A_{H11}$). 
    The collected reward for all the interactive approaches, including the one using the artificial trainer, presents a similar behavior showing a high correlation. 
    On the contrary, the collected reward by the autonomous agent shows a lower correlation in comparison to the interactive agents.
    }
    \label{fig:Correlation}
\end{figure}

Figure \ref{fig:Correlation} shows Pearson's correlation of the collected rewards for all the interactive agents trained by the participants in the experiment.
Moreover, we include an autonomous agent ($A_{Au}$) and an interactive agent trained by an artificial trainer agent ($A_{AT}$) as a reference.
It is possible to observe that all interactive agents, including the one using an artificial trainer agent, have a high correlation in terms of the collected reward.
However, the autonomous agent shows a lower correlation in comparison to the interactive approaches.

Additionally, we have computed the Student's t-test to test the statistical difference between the obtained results.
When the autonomous DeepRL approach is compared to agent-IDeepRL and human-IDeepRL, it obtains a t-score t=7.6829 (p-value p=\num{6.3947e-14}) and a t-score t=7.0192 (p-value p=\num{6.0755e-12}) respectively, indicating that there is a statistically significant difference between the two approaches.
On the other hand, comparing both interactive approaches between each other, i.e., agent-IDeepRL and human-IDeepRL, a t-score t=0.8461 (p-value p=0.3978) is obtained, showing that there is no statistical difference between the interactive methods.
Table \ref{Table:T-student} shows all the obtained t-scores along with the p-values for each of them.

\begin{table}[H]
\caption{Student's t-test for comparison of  autonomous DeepRL, agent-IDeepRL, and human-IDeepRL.}
\centering
\begin{tabular}{ccc}
\hline
\textbf{DeepRL vs.}	& \textbf{DeepRL vs.} & \textbf{agent-IDeepRL vs.}\\
\textbf{agent-IDeepRL} & \textbf{human-IDeepRL} & \textbf{human-IDeepRL}\\
\hline
$t = 7.6829$    & $t = 7.0192$  & $t = 0.8461$\\
$p = \num{6.3947e-14}$    & $p = \num{6.0755e-12}$  & $p=0.3978$\\
\hline
\end{tabular} \label{Table:T-student}
\end{table}


In all the tested approaches, approximately since episode 150, the agent performs actions mainly based on its learning or training.
In that episode, the value of $\epsilon$ in the $\epsilon$-greedy policy decay to 1\% of exploratory actions.
Moreover, in all the approaches, the maximal collected reward fluctuates between 2.5 and 3.
This is because the robot, with its movements, sometimes throws away another object from the table, different from the one being manipulated.

\section{Conclusions}

We have presented an interactive deep reinforcement learning approach to train an agent in a human-robot environment.
We have also performed a comparison between three different methods for learning agents.
First, we implemented an autonomous version of DeepRL, which had to interact and learn the environment by itself.
Next, we proposed an interactive version called IDeepRL, which used an external trainer to give useful advice during the decision-making process through interactive feedback delivered through early advising.

We have implemented two variations of IDeepRL by using previously trained artificial agents and humans as trainers.
We called these approaches agent-IDeepRL and human-IDeepRL, respectively.
Our proposed IDeepRL methods considerably outperform the autonomous DeepRL version in the implemented robotic scenario.
Moreover, in complex tasks, which often require more training time, to have an external trainer giving supportive feedback, leads to great benefits in terms of time and collected reward.

Overall, the interactive deep reinforcement learning approach introduces an advantage in domestic-like environments. 
It allows speeding up the learning process of a robotic agent interacting with the environment and allows people to transfer prior knowledge about a specific task. 
Furthermore, using a reinforcement learning approach allows the agent to learn the task without the necessity of previously labeled data, such as the case for supervised learning methods.
In this regard, the task is learned in such a way that the agent learns to recognize the state of the environment as well as to behave on it, deciding where to place the different objects. 
Our novel interactive-shaping vision-based approach outperforms the current autonomous DeepRL method, used as a baseline in this work. 
The introduced approaches demonstrate that the use of external trainers, either artificial or human, lead to more and faster reward in comparison to traditional the DeepRL approach. 

As future work, we consider the use of different kinds of artificial trainers to select possible advisors better.
A bad teacher can negatively influence the learning process and somehow limit the learner by teaching a strategy that is not necessarily optimal.
To select a good teacher, it is necessary to take into account that an agent that obtains the best results for the task, in terms of accumulated reward, is not necessarily the best teacher \cite{cruz2018improving}.
Rather a good teacher could be one with a small standard deviation over the visited states. 
This would allow advising the learner agent in more specific situations.
Additionally, we plan to transfer and test the proposed approach in a real-world human-robot interaction scenario.
In such a case, it is necessary to deal with additional environmental dynamics and variables that in simulated scenarios may be easier to control.




\section*{Acknowledgments}
This research was partially funded by Universidad Central de Chile under the research project CIP2018009, the Coordenaç\~ao de Aperfeiçoamento de Pessoal de N\'ivel Superior - Brasil (CAPES) - Finance Code 001, the Brazilian agencies FACEPE, and CNPq.





\bibliographystyle{ieeetr}

\bibliography{paper}

\begin{thebibliography}{10}

\bibitem{shepherd2014kuka}
S.~Shepherd and A.~Buchstab, ``Kuka robots on-site,'' in {\em Robotic
  Fabrication in Architecture, Art and Design 2014}, pp.~373--380, Springer,
  2014.

\bibitem{cruz2018action}
F.~Cruz, P.~W{\"u}ppen, A.~Fazrie, C.~Weber, and S.~Wermter, ``Action selection
  methods in a robotic reinforcement learning scenario,'' in {\em 2018 IEEE
  Latin American Conference on Computational Intelligence (LA-CCI)}, pp.~1--6,
  IEEE, 2018.

\bibitem{goodrich2008human}
M.~A. Goodrich, A.~C. Schultz, {\em et~al.}, ``Human--robot interaction: a
  survey,'' {\em Foundations and Trends{\textregistered} in Human--Computer
  Interaction}, vol.~1, no.~3, pp.~203--275, 2008.

\bibitem{churamani2020icub}
N.~Churamani, F.~Cruz, S.~Griffiths, and P.~Barros, ``icub: learning emotion
  expressions using human reward,'' {\em arXiv preprint arXiv:2003.13483},
  2020.

\bibitem{sutton2018reinforcement}
R.~S. Sutton and A.~G. Barto, {\em Reinforcement learning: {A}n introduction}.
\newblock MIT press, 2018.

\bibitem{niv2009reinforcement}
Y.~Niv, ``Reinforcement learning in the brain,'' {\em Journal of Mathematical
  Psychology}, vol.~53, no.~3, pp.~139--154, 2009.

\bibitem{cruz2018multi}
F.~Cruz, G.~I. Parisi, and S.~Wermter, ``Multi-modal feedback for
  affordance-driven interactive reinforcement learning,'' in {\em 2018
  International Joint Conference on Neural Networks (IJCNN)}, pp.~1--8, IEEE,
  2018.

\bibitem{bignold2020conceptual}
A.~Bignold, F.~Cruz, M.~E. Taylor, T.~Brys, R.~Dazeley, P.~Vamplew, and
  C.~Foale, ``A conceptual framework for externally-influenced agents: An
  assisted reinforcement learning review,'' {\em arXiv preprint
  arXiv:2007.01544}, 2020.

\bibitem{ayala2019reinforcement}
A.~Ayala, C.~Henr{\'\i}quez, and F.~Cruz, ``Reinforcement learning using
  continuous states and interactive feedback,'' in {\em Proceedings of the
  International Conference on Applications of Intelligent Systems}, pp.~1--5,
  2019.

\bibitem{millan2019human}
C.~Mill{\'a}n, B.~Fernandes, and F.~Cruz, ``Human feedback in continuous
  actor-critic reinforcement learning,'' in {\em Proceedings of the European
  Symposium on Artificial Neural Networks, Computational Intelligence and
  Machine Learning ESANN}, pp.~661--666, ESANN, 2019.

\bibitem{barros2020moody}
P.~Barros, A.~Tanevska, F.~Cruz, and A.~Sciutti, ``Moody learners -- explaining
  competitive behaviour of reinforcement learning agents,'' in {\em Proceedings
  of IEEE International Conference on Development and Learning and Epigenetic
  Robotics (ICDL-EpiRob)}, 2020.

\bibitem{lecun2015deep}
Y.~LeCun, Y.~Bengio, and G.~Hinton, ``Deep learning,'' {\em nature}, vol.~521,
  no.~7553, pp.~436--444, 2015.

\bibitem{mnih2015human}
V.~Mnih, K.~Kavukcuoglu, D.~Silver, A.~A. Rusu, J.~Veness, M.~G. Bellemare,
  A.~Graves, M.~Riedmiller, A.~K. Fidjeland, G.~Ostrovski, {\em et~al.},
  ``Human-level control through deep reinforcement learning,'' {\em Nature},
  vol.~518, no.~7540, p.~529, 2015.

\bibitem{krizhevsky2012imagenet}
A.~Krizhevsky, I.~Sutskever, and G.~E. Hinton, ``Imagenet classification with
  deep convolutional neural networks,'' in {\em Advances in neural information
  processing systems}, pp.~1097--1105, 2012.

\bibitem{van2016deep}
H.~Van~Hasselt, A.~Guez, and D.~Silver, ``Deep reinforcement learning with
  double q-learning,'' in {\em Thirtieth AAAI conference on artificial
  intelligence}, pp.~2094--2100, 2016.

\bibitem{puterman1994markov}
M.~L. Puterman, {\em Markov Decision Processes: Discrete Stochastic Dynamic
  Programming}.
\newblock Wiley, 1994.

\bibitem{suay2011effect}
H.~B. Suay and S.~Chernova, ``Effect of human guidance and state space size on
  interactive reinforcement learning,'' in {\em 2011 Ro-Man}, pp.~1--6, IEEE,
  2011.

\bibitem{najar2020reinforcement}
A.~Najar and M.~Chetouani, ``Reinforcement learning with human advice. a
  survey,'' {\em arXiv preprint arXiv:2005.11016}, 2020.

\bibitem{ng1999policy}
A.~Y. Ng, D.~Harada, and S.~Russell, ``Policy invariance under reward
  transformations: Theory and application to reward shaping,'' in {\em ICML},
  vol.~99, pp.~278--287, 1999.

\bibitem{brys2014combining}
T.~Brys, A.~Now{\'e}, D.~Kudenko, and M.~E. Taylor, ``Combining multiple
  correlated reward and shaping signals by measuring confidence.,'' in {\em
  AAAI}, pp.~1687--1693, 2014.

\bibitem{griffith2013policy}
S.~Griffith, K.~Subramanian, J.~Scholz, C.~Isbell, and A.~L. Thomaz, ``Policy
  shaping: Integrating human feedback with reinforcement learning,'' in {\em
  Advances in Neural Information Processing Systems}, pp.~2625--2633, 2013.

\bibitem{li2019human}
G.~Li, R.~Gomez, K.~Nakamura, and B.~He, ``Human-centered reinforcement
  learning: A survey,'' {\em IEEE Transactions on Human-Machine Systems},
  vol.~49, no.~4, pp.~337--349, 2019.

\bibitem{grizou2013robot}
J.~Grizou, M.~Lopes, and P.-Y. Oudeyer, ``Robot learning simultaneously a task
  and how to interpret human instructions,'' in {\em 2013 IEEE Third Joint
  International Conference on Development and Learning and Epigenetic Robotics
  (ICDL)}, pp.~1--8, IEEE, 2013.

\bibitem{navidi2020human}
N.~Navidi, ``Human ai interaction loop training: New approach for interactive
  reinforcement learning,'' {\em arXiv preprint arXiv:2003.04203}, 2020.

\bibitem{bignold2019rule}
A.~Bignold, {\em Rule-based interactive assisted reinforcement learning}.
\newblock PhD thesis, Federation University Australia, 2019.

\bibitem{taylor2014reinforcement}
M.~E. Taylor, N.~Carboni, A.~Fachantidis, I.~Vlahavas, and L.~Torrey,
  ``Reinforcement learning agents providing advice in complex video games,''
  {\em Connection Science}, vol.~26, no.~1, pp.~45--63, 2014.

\bibitem{cruz2018improving}
F.~Cruz, S.~Magg, Y.~Nagai, and S.~Wermter, ``Improving interactive
  reinforcement learning: What makes a good teacher?,'' {\em Connection
  Science}, vol.~30, no.~3, pp.~306--325, 2018.

\bibitem{dobrovsky2019improving}
A.~Dobrovsky, U.~M. Borghoff, and M.~Hofmann, ``Improving adaptive gameplay in
  serious games through interactive deep reinforcement learning,'' in {\em
  Cognitive infocommunications, theory and applications}, pp.~411--432,
  Springer, 2019.

\bibitem{rajeswaran2017learning}
A.~Rajeswaran, V.~Kumar, A.~Gupta, G.~Vezzani, J.~Schulman, E.~Todorov, and
  S.~Levine, ``Learning complex dexterous manipulation with deep reinforcement
  learning and demonstrations,'' {\em arXiv preprint arXiv:1709.10087}, 2017.

\bibitem{lukka2014zenrobotics}
T.~J. Lukka, T.~Tossavainen, J.~V. Kujala, and T.~Raiko, ``Zenrobotics
  recycler--robotic sorting using machine learning,'' in {\em Proceedings of
  the International Conference on Sensor-Based Sorting (SBS)}, 2014.

\bibitem{zhihong2017robotgrasping}
C.~{Zhihong}, Z.~{Hebin}, W.~{Yanbo}, L.~{Binyan}, and L.~{Yu}, ``A
  vision-based robotic grasping system using deep learning for garbage
  sorting,'' in {\em 2017 36th Chinese Control Conference (CCC)},
  pp.~11223--11226, 2017.

\bibitem{sun2017clothing}
L.~{Sun}, G.~{Aragon-Camarasa}, S.~{Rogers}, R.~{Stolkin}, and J.~P. {Siebert},
  ``Single-shot clothing category recognition in free-configurations with
  application to autonomous clothes sorting,'' in {\em 2017 IEEE/RSJ
  International Conference on Intelligent Robots and Systems (IROS)},
  pp.~6699--6706, 2017.

\bibitem{cruz2014improving}
F.~Cruz, S.~Magg, C.~Weber, and S.~Wermter, ``Improving reinforcement learning
  with interactive feedback and affordances,'' in {\em 4th International
  Conference on Development and Learning and on Epigenetic Robotics},
  pp.~165--170, IEEE, 2014.

\bibitem{cruz2016learning}
F.~Cruz, G.~I. Parisi, and S.~Wermter, ``Learning contextual affordances with
  an associative neural architecture,'' in {\em Proceedings of the European
  Symposium on Artificial Neural Network. Computational Intelligence and
  Machine Learning ESANN}, pp.~665--670, UCLouvain, 2016.

\bibitem{zhang2015towards}
F.~Zhang, J.~Leitner, M.~Milford, B.~Upcroft, and P.~Corke, ``Towards
  vision-based deep reinforcement learning for robotic motion control,'' {\em
  arXiv preprint arXiv:1511.03791}, 2015.

\bibitem{vecerik2017leveraging}
M.~Vecerik, T.~Hester, J.~Scholz, F.~Wang, O.~Pietquin, B.~Piot, N.~Heess,
  T.~Roth{\"o}rl, T.~Lampe, and M.~Riedmiller, ``Leveraging demonstrations for
  deep reinforcement learning on robotics problems with sparse rewards,'' {\em
  arXiv preprint arXiv:1707.08817}, 2017.

\bibitem{desai2017deep}
N.~Desai and A.~Banerjee, ``Deep reinforcement learning to play space
  invaders,'' tech. rep., Stanford University, 2017.

\bibitem{adam2012experience}
S.~Adam, L.~Busoniu, and R.~Babuska, ``Experience replay for real-time
  reinforcement learning control,'' {\em Systems, Man, and Cybernetics, Part C:
  Applications and Reviews, IEEE Transactions on}, vol.~42, pp.~201 -- 212, 04
  2012.

\bibitem{cruz2017agent}
F.~Cruz, P.~W{\"u}ppen, S.~Magg, A.~Fazrie, and S.~Wermter, ``Agent-advising
  approaches in an interactive reinforcement learning scenario,'' in {\em 2017
  Joint IEEE International Conference on Development and Learning and
  Epigenetic Robotics (ICDL-EpiRob)}, pp.~209--214, IEEE, 2017.

\bibitem{rohmer2013vrep}
E.~Rohmer, S.~P. Singh, and M.~Freese, ``V-rep: A versatile and scalable robot
  simulation framework,'' in {\em IEEE/RSJ International Conference on
  Intelligent Robots and Systems}, pp.~1321--1326, IEEE, 2013.

\end{thebibliography}

\end{document}